\documentclass{article}
\usepackage{spconf,amsmath,graphicx}
\usepackage{multirow}
\usepackage{indentfirst}
\graphicspath{{image_pdf/}}
\usepackage{adjustbox}
\usepackage{lipsum}
\usepackage{amsmath}
\usepackage{boxedminipage}
\usepackage{scrextend}
\usepackage[left=1.5cm,top=2cm,right=1.5cm,bottom=1.5cm]{geometry}
\usepackage[table,usenames,dvipsnames]{xcolor}
\usepackage{tikz}
\usepackage{subfig} 
\usepackage{pgfplots}
\usepackage{caption}
\pgfplotsset{
    width=\textwidth,
    ylabel right/.style={
        after end axis/.append code={
            \node [rotate=90, anchor=north] at (rel axis cs:1,0.5) {#1};
        }   
    }
}
\title{Fixed-point optimization of deep neural networks \\with adaptive step size retraining}
\name{Sungho Shin, Yoonho Boo and Wonyong Sung\thanks{This work was supported in part by the Brain Korea 21 Plus Project and the National Research Foundation of Korea (NRF) grant funded by the Korea government (MSIP) (No. 2015R1A2A1A10056051).}}
\address{ Department of Electrical and Computer Engineering\\
 Seoul National University\\
  Seoul, 08826 Korea \\
Email: sungho.develop@gmail.com, yhboo.research@gmail.com, wysung@snu.ac.kr \\}
\begin{document}
%
\maketitle
\begin{abstract}
Fixed-point optimization of deep neural networks plays an important role in hardware based design and low-power implementations. Many deep neural networks show fairly good performance even with 2- or 3-bit precision when quantized weights are fine-tuned by retraining. We propose an improved fixed-point optimization algorithm that estimates the quantization step size dynamically during the retraining. In addition, a gradual quantization scheme is also tested, which sequentially applies fixed-point optimizations from high- to low-precision. The experiments are conducted for feed-forward deep neural networks (FFDNNs), convolutional neural networks (CNNs), and recurrent neural networks (RNNs).
\end{abstract}
\begin{keywords}
deep neural networks, recurrent neural networks, fixed-point quantization, step size adaptation
\end{keywords}
\section{Introduction}
\label{sec:1}
Deep neural networks (DNNs) show very high performance in various fields such as speech recognition~\cite{amodei2015deep} and image classification~\cite{He_2016_CVPR}. However, real-time implementation of DNNs usually demands many arithmetic and weight fetch operations. Thus, word-length optimization is needed in embedded applications to reduce the strength of arithmetic and the size of the weight storage. However, reducing the word length too much tends to degrade the performance. Thus, developing optimum quantization methods is greatly needed for efficient implementation of neural network algorithms. 

Direct quantization of deep neural networks usually does not show satisfactory performance with very low precision weights. However, when the quantized weights are optimized by retraining, the fixed-point performance improves dramatically. Even ternary valued weights (+1, 0, and -1) for a DNN have yielded satisfactory performance~\cite{shin2016fixed, hwang2014fixed}. Recently, several improved fixed-point optimization methods are developed by employing retraining based fine tuning~\cite{lin2015fixed, han2015deep}. Also, VLSI and FPGA based deep neural networks have been implemented using fixed-point weights~\cite{kim2014x1000, park2016fpga, HanLMPPHD16, lee2016fpga, fraser2017scaling}.
 
In this work, an improved retraining algorithm is developed for fixed-point optimization of deep neural networks. The previous works decide the optimum quantization step size based on the distribution of floating-point weights and freezes the step-size during the retraining period~\cite{hwang2014fixed, lin2015fixed}. The proposed algorithm adaptively determines the step-size at the re-quantization step during retraining. Since the weight values change much at the beginning of retraining, this approach is especially effective when applied at initial retraining epochs. In order to change the weight values less abruptly, we also propose and evaluate the gradual quantization method.  In this schemes, floating-point weights are converted to, for example, 6-bit weights, which are then converted to 4-bit weights, and so on. We evaluate the proposed schemes in three different networks: feed-forward deep neural networks (FFDNNs), convolutional neural networks (CNNs), and recurrent neural networks (RNNs). The proposed methods yielded better results compared to the previous retrain-based quantization schemes. 

The rest of this paper is organized as follows. Section~\ref{sec:3} presents the proposed quantization with step size adaptation during the retraining procedure. The gradual quantization scheme is also explained. Experimental results on FFDNN, CNN, and RNN applications are shown in Section~\ref{sec:4}. Concluding remarks follow in Section~\ref{sec:5}.
\begin{figure}[t]
\begin{center}
	\begin{boxedminipage}{0.48\textwidth}\bfseries\fontsize{9}{5}
		\raggedright \fontsize{9}{5}\selectfont{- Quantization step size determining:}
		\center{$\Delta = QStep(\boldsymbol{w})=  \operatornamewithlimits{argmin}\limits_{{\Delta}}\operatorname{}\dfrac1 2\sum\limits_{i=1}^N\bigl(Q(w_{i},\Delta)-w_{i} \bigr)^2$ \\}
		\raggedright - Quantized weights:
		\center{$	\boldsymbol{w}^{(q)}=Q(\boldsymbol{w},\Delta)= sgn(\boldsymbol{w}) \cdot \Delta  \cdot min\biggl(\left \lfloor{\dfrac{\lvert \boldsymbol{w} \rvert}\Delta +0.5}\right \rfloor ,\dfrac{M-1}2\biggr)$ \\}
		\raggedright - Forward:
		\center{$net_{i} = \sum\limits_{j \in A_{i}}w_{ij}^{(q)}y_{j}$ \\}
		\center{$y_{i}=\phi_{i}(net_{i})$ \\}
		\raggedright - Backward:
		\center{$\delta_{j}=\phi_{j}^{'}(net_{j})\sum\limits_{i \in P_{j}}\delta_{i}w_{ij}^{(q)}$ \\}
		\raggedright - Gradient calculation:
		\center{$\dfrac {\partial E}{\partial w_{ij}} = -\delta_{i}y_{j}$ \\}
		\raggedright - Weights update:
		\center$w_{ij, new} = w_{ij} - \alpha  \dfrac {\partial E}{\partial w_{ij}}$ \\
		\center{$\Delta_{new}=QStep(w_{ij, new})$}   \raggedleft \fontsize{9}{5}\selectfont(Proposed scheme) \fontsize{9}{5}
		\center$w_{ij, new}^{(q)} = Q_{ij}(w_{ij, new},\Delta_{new})$ \\
	\end{boxedminipage}
\end{center}
	\caption{Overall fixed-point retraining algorithm with step size adaptation scheme, where $\Delta$ is the quantization step size, $\boldsymbol{w}$ is the weight groups, $net_i$ is the summed input value of unit $i$, $\delta_i$ is the error signal of unit $i$, $M$ is quantization points (2-bit quantization = 3 points, 3-bit quantization = 7 points), $\alpha$ is the learning rate, $N$ is the number of the weights in each layer, $A_{i}$ and $P_{j}$ represent the activation of next and previous layer, $\phi(\cdot)$ is the activation function, $E$ is the output error, and superscript $(q)$ means the value is quantized.}
\label{fig1}
\end{figure}
\section{Step size adaptation AND GRADUAL QUANTIZATION for retraining of deep neural networks}
\label{sec:3}
In this section, we explain the conventional retrain based fixed-point optimization algorithm, and present adaptive step size retraining and gradual quantization methods.  

\subsection{Retrain-based fixed-point quantization review}
\label{ssec:remind}
The original retrain based fixed-point optimization algorithm can be represented briefly as shown in~\figurename~\ref{fig1}. Note that, conventional algorithms~\cite{shin2016fixed, hwang2014fixed, anwar2015fixed} do not compute $\Delta_{new}$ at the `weights update' stage. In this figure, after obtaining the floating-point weights by training, the quantization step size, $\Delta$, is determined by minimizing the L2 error between the floating-point and fixed-point weights. For the convenience of arithmetic, uniform quantization is assumed. Two algorithms have been developed for the quantization step size optimization. One is an exhaustive search, which decides the initial quantization step size $\Delta_{initial}$ by considering the weight distribution, and then searches the best performing step size between $\Delta_{initial}/2$ and $2\cdot\Delta_{initial}$ by testing the quantized network with the evaluation set~\cite{hwang2014fixed}. The second approach is deciding the quantization step size by measuring the mean and the variance of the floating-point weights~\cite{lin2015fixed}. Then, in the second stage of~\figurename~\ref{fig1}, floating-point weights are rounded to fixed-point values by using the determined quantization step size. The third stage is the inferencing or the forward stage with the quantized network, $w^{(q)}$. The error signal is calculated and used for backward propagation.  The gradient is calculated and weight update is conducted. Note that the floating-point weights, instead of the fixed-point values, are updated because the amount of weight update is usually much smaller than the quantization step size. Then, the fixed-point weight update, yielding $w_{ij, new}^{(q)}$, is accomplished by quantizing the updated floating-point weights. Note that determining $\Delta_{new}$ is not performed in the conventional method, and the same quantization step size is used at every iteration.  

\subsection{Step-size adaptation during retraining}
\label{ssec:stepsize}
\begin{table*}[]
\centering
\caption{Frame-level phoneme error rate (\%) on the test set with the TIMIT phoneme recognition examples. Note that `conventional' is the baseline~\cite{hwang2014fixed} and `adaptive' is the proposed scheme.}
\label{table:dnn}
\begin{tabular}{cc|ccccc|ccccc}
\hline\hline
                                                                            &                    & \multicolumn{5}{c|}{Without BN}                                                    & \multicolumn{5}{c}{With BN}                                                       \\ \hline
                                                                            & Size of each layer & 64             & 128            & 256            & 512            & 1024           & 64             & 128            & 256            & 512            & 1024           \\ \hline
                                                                            & Floating result    & \textbf{34.38} & \textbf{31.63} & \textbf{30.17} & \textbf{29.61} & \textbf{29.53} & \textbf{33.82} & \textbf{30.81} & \textbf{29.79} & \textbf{29.77} & \textbf{29.59} \\ \hline\hline
\multirow{3}{*}{\begin{tabular}[c]{@{}c@{}}2-bit\\ (3 point)\end{tabular}}  & Direct             & 80.25          & 84.12          & 81.92          & 83.30          & 75.05          & 89.82          & 88.79          & 87.57          & 85.73          & 86.10          \\ 
                                                                            & Conventional       & 43.73          & 37.80          & 33.70          & 31.43          & 29.99          & 41.81          & 35.88          & 33.12          & 31.21          & 30.22          \\  
                                                                            & Adaptive           & \textbf{42.06} & \textbf{36.88} & \textbf{32.61} & \textbf{30.61} & \textbf{29.49} & \textbf{37.87} & \textbf{33.46} & \textbf{31.48} & \textbf{30.73} & \textbf{30.09} \\ \hline\hline
\multirow{3}{*}{\begin{tabular}[c]{@{}c@{}}3-bit\\ (7 point)\end{tabular}}  & Direct             & 68.13          & 63.65          & 60.33          & 51.46          & 48.61          & 80.41          & 69.55          & 69.42          & 81.60          & 64.55          \\  
                                                                            & Conventional       & 40.63          & 34.73          & 31.41          & 30.49          & \textbf{29.33} & 36.88          & 32.58          & 30.53          & 30.14          & 29.76          \\ 
                                                                            & Adaptive           & \textbf{37.89} & \textbf{33.80} & \textbf{30.74} & \textbf{29.83} & 29.40          & \textbf{35.29} & \textbf{31.94} & \textbf{30.32} & \textbf{30.10} & \textbf{29.65} \\ \hline\hline
\multirow{3}{*}{\begin{tabular}[c]{@{}c@{}}4-bit\\ (15 point)\end{tabular}} & Direct             & 58.90          & 50.58          & 42.15          & 38.05          & 36.53          & 65.63          & 50.43          & 46.46          & 43.80          & 39.77          \\ 
                                                                            & Conventional       & 36.51          & 32.65          & 30.79          & 29.95          & 29.44          & 34.17          & 31.34          & 29.86          & \textbf{29.81} & 29.70          \\ 
                                                                            & Adaptive           & \textbf{35.50} & \textbf{32.09} & \textbf{30.50} & \textbf{29.54} & \textbf{29.29} & \textbf{33.91} & \textbf{30.86} & \textbf{29.47} & 29.87          & \textbf{29.52} \\ \hline\hline
\end{tabular}
\end{table*}

As described in Section~\ref{ssec:remind}, the conventional method freezes the step size during the retraining. However, in many cases, the weight values change much by retraining. Note that the amount of change decreases as the retraining iteration progresses. Thus, it is advantageous for improving the performance to adjust the quantization step size during the retraining. Especially, the need of step size adaptation is greater at the beginning of retraining. The proposed scheme adds the determination of $\Delta_{new}$ at the weight update stage of~\figurename~\ref{fig1}. 

We do not perform `exhaustive search' anymore but update the quantization step size during retraining by using the L2 error minimization between the floating-point and fixed-point weights. We consider two different quantization step size update timing. The first one is `epoch-level update', and the other is `1 epoch update \& fix'. The `epoch-level update' changes the step size at every epoch. The `1 epoch update \& fix' updates the step size only during one or two epochs and freezes it for the remaining epochs. In our empirical evaluation, the first scheme is good for FFDNNs, but the second one shows better results  for CNNs and RNNs. The specific results will be given in Section~\ref{sec:4}.

\subsection{Gradual quantization scheme}
\label{ssec:gradual}
We also propose another step size adaptation approach which is similar to the curriculum learning. The curriculum learning is a training strategy to move the goal from an easy level to more complex one gradually~\cite{bengio2009curriculum}. One of the important points in curriculum learning is how to organize the tasks from easy to complex ones. We consider that the fixed-point optimization with a small number of bits is a more difficult problem than that with a large one. 

In the proposed scheme, we begin fixed-point optimization with a fairly high precision, such as 6 bits, and then keep lowering the word-length by one bit with retraining for each precision. At each retraining process with a given precision, we also combine the proposed quantization step size adaptation scheme. The experiments are conducted for FFDNNs.

\section{Experimental Results}
\label{sec:4}
The proposed step size adaptation is evaluated for three applications. We employ FFDNNs for phoneme recognition, CNNs for house number recognition, and RNNs for language modeling. To analyze the effect of step size adaptation, we change the size of each network and their word lengths. 
\begin{figure}[t]
\centering
\centerline{
\begin{tikzpicture}
    \begin{axis}[
	width=\columnwidth,
	height = 0.5\columnwidth,
	compat=1.12,
	xmin=0,
	ymin=0.05,
	xmax=19,
	ymax=0.5,
	label style={font=\footnotesize},
	legend style={font=\footnotesize,at={(0.05,0.95)},anchor=north west},
	tick label style={font=\scriptsize}, 
	domain=1:512, 
	minor x tick num=4, 
	minor y tick num=4, 
	log basis x={10}, 
	xtick pos=both, 
	xtick align=inside, 
	major tick style={line width=0.010cm, black},
	 major tick length=0.10cm,
        xlabel=Number of epochs,
        ylabel=$\Delta_{adapt}$,
	xlabel shift=-3pt,
	ylabel shift=-3pt]
	\legend{Layer1, Layer2, Layer3};
	\addplot[color=blue, mark=x, mark size=1.6pt, solid, mark repeat=1,mark options=solid] file{data/curve_256_l1.txt}; 
	\addplot[color=yellow!60!black, mark=otimes, mark size=1.6pt, solid, mark repeat=1,mark options=solid] file{data/curve_256_l2.txt};
	\addplot[color=green, mark=square, mark size=1.6pt, solid, mark repeat=1,mark options=solid] file{data/curve_256_l3.txt}; 
    \end{axis}
   \end{tikzpicture}}
\caption{Training curves in terms of $\Delta_{adapt}$ with the 256 size of the FFDNN.} 
\label{fig:2}
\end{figure}
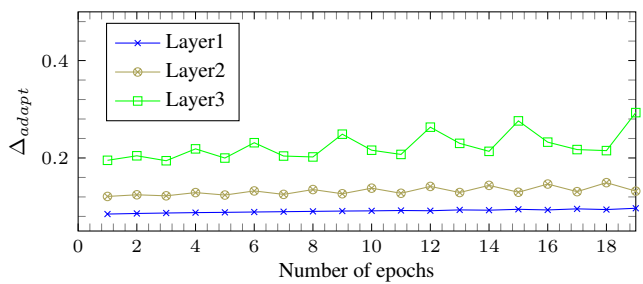

\begin{table}[]
\centering
\caption{Evaluation of the proposed quantization strategies on TIMIT phoneme recognition task. The network is FFDNN with two 512 size hidden layers, and the floating-point result is 29.61\%. `Conventional' is general retraining based quantization, `adaptive' conducts proposed step size adaptation, `gradual' is curriculum learning style quantization scheme, and `adaptive \& gradual' represents mixed approach using both techniques.}
\label{table:grad}
\begin{tabular}{ccccc}
\hline\hline
      & Conventional & Adaptive & Gradual & \begin{tabular}[c]{@{}c@{}}Adaptive \\ \& \\ Gradual\end{tabular} \\ \hline\hline
6-bit & \textbf{29.32}         & \textbf{29.32}      & \textbf{29.32}   & \textbf{29.32}                                                               \\ \hline
4-bit & 29.95         & 29.54       & 29.53   & \textbf{29.49}                                                               \\ \hline
3-bit & 30.49         & 29.83      & 29.90   & \textbf{29.61}                                                               \\ \hline
2-bit & 31.43         & \textbf{30.61}      & 30.69   & 30.62                                                               \\ \hline\hline
\end{tabular}
\end{table}
\subsection{Phoneme recognition using feed-forward deep neural networks}
\label{ssec:dnn}

The FFDNN is trained with the TIMIT corpus~\cite{garofolo1993darpa}, and the detailed experimental condition for the data preprocessing is the same with~\cite{graves2013speech}. We construct 11 consecutive frames as the network input. The output layer supports 61 labels, and the labels are merged into 39 classes for the final evaluation. For performance evaluation, the number of units in each layer increases from 64 to 1024. We train the floating-point networks using the stochastic gradient descent (SGD) with Nesterov momentum~\cite{sutskever2013importance}. The learning rate decreases from 2e-3 to 3.90625e-6 with a factor of 2 when the development set does not show improvements for 4 consecutive evaluations. For fixed-point networks training, all other conditions are the same with the floating-point case but the initial learning rate is 5e-4.

The results of fixed-point optimization for FFDNNs with and without the step size adaptation are reported in~\tablename~\ref{table:dnn}. The experiments also show the results with batch normalization~(BN)~\cite{43442}. The step size is updated using the `epoch-level update' until the end of the retraining. \tablename~\ref{table:dnn} shows that the floating-point network performance saturates at 512 units size when BN is applied, and at 256 units when BN is not used. When the unit size in each layer is 512 or smaller, the proposed algorithm yields better performance in both cases. For example, if the 512 units size network is quantized in 2-bit without BN, the differences between the floating-point and the fixed-point networks are 1.82\% and 1\% for `conventional' and `adaptive' schemes, respectively. In addition, the phoneme error rate of the 3-bit network optimized with the `adaptive' scheme (29.83\%) is lower than that of the 4-bit quantized network with the `conventional' scheme (29.95\%). 

BN improves the performance of both floating-point and fixed-point networks. Applying the `adaptive' method improves the performance. For example, if the layer unit size is 128 and 2-bit quantization is used, BN brings the performance gain of 3.42\% when `adaptive' scheme is used. Therefore, the proposed `adaptive' method can be efficiently used with BN.

When the unit size is large enough, the quantization scheme does not affect the performance much because a larger size network has a better resiliency to quantization~\cite{sung2015resiliency}. Even the 4-bit quantized 512 units size network without BN shows the performance almost comparable to the floating-point 1024 units size network. When the network is trained with BN, it shows a similar trend.

\figurename~\ref{fig:2} shows the step size $\Delta$ of the proposed adaptive scheme as the retraining progresses. Note that the step size is renewed at each epoch during retraining. As shown in this figure, the step size of the last layer varies much, while that of the first layer is almost constant. The step size adaptation is much needed for the last layer. 

We also evaluate the performance of the gradual quantization scheme. The results are reported in~\tablename~\ref{table:grad}. The floating point results show 29.61\% error rate on the test set. The 6-bit word length shows slightly better accuracy than the floating point. Thus, we define the easiest task as the 6-bit quantization. In~\tablename~\ref{table:grad}, the `gradual' scheme yields better performance than the `conventional' strategy, but shows worse or similar results compared to the `adaptive' quantization. The combined strategy of the `adaptive' and `gradual' shows slightly better accuracy than the `adaptive' strategy in 4- and 3-bit quantization, but it is worse than the `adaptive' scheme in 2-bit quantization. Since there is no performance difference between the `adaptive' and `adaptive \& gradual' scheme, we only employ the 'adaptive' scheme for CNN and RNN experiments.

\subsection{Image classification using convolutional neural networks}
\label{ssec:cnn}
\begin{table}[]
\centering
\caption{Miss classification rate on the test set with the SVHN house number recognition example. The alphabets `L', `C', and `V' represent specific structure of the CNN. The `L' is the most smallest network and the `V' is the biggest network. Please refer Section~\ref{ssec:cnn} for details.}
\label{table:cnn}
\begin{tabular}{ccccc}
\hline\hline
                                                                                     & {Type of network} & {L}       & {C}        & {V}        \\ \hline
                                                                                     & {Floating result} & \textit{\textbf{6.45}} & \textit{\textbf{5.65}} & \textit{\textbf{4.50}} \\ \hline\hline
\multirow{3}{*}{{\begin{tabular}[c]{@{}c@{}}2-bit\\ (3 point)\end{tabular}}}  & {Direct}          & 45.68            & 23.17             &  73.55            \\ 
                                                                                     & {Conventional}   & 8.37            & 7.10              & {5.24}     \\ 
                                                                                     & {Adaptive}      & \textbf{8.01}    & \textbf{6.65}     &     \textbf{5.02}          \\ \hline\hline
\multirow{3}{*}{{\begin{tabular}[c]{@{}c@{}}3-bit\\ (7 point)\end{tabular}}}  & {Direct}          & 10.14            & 7.88              &  6.73             \\
                                                                                     & {Conventional}   & 7.04             & 5.97              & {4.57}     \\  
                                                                                     & {Adaptive}      & \textbf{6.92}    & \textbf{5.91}     &    \textbf{4.53}          \\ \hline\hline
\multirow{3}{*}{{\begin{tabular}[c]{@{}c@{}}4-bit\\ (15 point)\end{tabular}}} & {Direct}          & 7.85             & 6.03              &   4.79            \\ 
                                                                                     & {Conventional}   & {6.60}    & \textbf{5.76}     &    4.74           \\
                                                                                     & {Adaptive}      &\textbf{6.46}            & 5.86              & \textbf{4.60}     \\ \hline
\end{tabular}
\end{table}
\begin{table}[t]
\centering
\caption{Bit per character (BPC) on the test set with the English Wikipedia language model. }
\label{table:rnn}
\begin{tabular}{ccccc}
\hline\hline
\textbf{}                                                                   & Size of each layer      & 64                      & 128                     & 256                     \\ \hline
\textbf{}                                                                   & Floating result         & \textit{\textbf{2.07}} & \textit{\textbf{1.81}} & \textit{\textbf{1.65}} \\ \hline\hline
\multirow{3}{*}{\begin{tabular}[c]{@{}c@{}}2-bit\\ (3 point)\end{tabular}}  & Direct                  & 8.46                   & 9.53                   & 7.26                   \\  
                                                                            & Conventional           & 2.48                   & 2.49                    & 1.89                   \\  
                                                                            &Adaptive& \textbf{2.42}          & \textbf{2.16}          & \textbf{1.86}          \\ \hline\hline
\multirow{3}{*}{\begin{tabular}[c]{@{}c@{}}3-bit\\ (7 point)\end{tabular}}  & Direct                  & 7.176                   & 6.84                   & 4.35                   \\ 
                                                                            & Conventional           & 2.52                   & 2.10                   & 1.91                   \\  
                                                                            &Adaptive & \textbf{2.35}          & \textbf{2.06}          & \textbf{1.82}          \\ \hline\hline
\multirow{3}{*}{\begin{tabular}[c]{@{}c@{}}4-bit\\ (15 point)\end{tabular}} & Direct                  & 4.49                   & 5.50                   & 2.59                   \\  
                                                                            & Conventional           & 2.43                   & 2.04                   & \textbf{1.83}          \\  
                                                                            & Adaptive & \textbf{2.32}          & \textbf{1.95}          & 1.86                   \\ \hline\hline
\multirow{3}{*}{\begin{tabular}[c]{@{}c@{}}6-bit\\ (63 point)\end{tabular}} & Direct                  & 2.56                   &   3.73                 &   1.73      \\  
                                                                            & Conventional           &\textbf{2.11}                  &  \textbf{1.87}                 & \textbf{1.67}          \\  
                                                                            & Adaptive & \textbf{2.11}          & {1.89}          &          1.68         \\ \hline\hline
\end{tabular}
\end{table}
Image classification experiments are performed on the SVHN dataset~\cite{netzer2011reading}. The dataset includes 600,000 labeled 32x32 three channel images from real world house numbers. For the data preprocessing, we employ the same method with~\cite{sermanet2012convolutional}. The output label has ten units which represent the numbers from 0 to 9. For evaluation of the proposed scheme, we employ three different structures. We name the networks as `L', `C', and `V' which have the trainable parameters 60k, 84k, and 435k, respectively. The `L' network is Lenet5~\cite{lecun1989backpropagation}, `C' network is from~\cite{krizhevsky2012cuda}, and `V' network is constructed  as VGG style which is from~\cite{courbariaux2015binaryconnect}. We train the floating-point networks using SGD with Nesterov momentum. The learning rate is decreased from 2-e2 to 3.125e-4 with a factor of 2 when the development set does not show improvement for 4 consecutive evaluations. For the fixed-point network training, the initial learning rate was 5e-4. The effects of step size adaptation in the CNNs are examined in~\tablename~\ref{table:cnn}. The step size is updated using the `1 epoch update \& fix' strategy. Our algorithm works well for `L' and `V' networks regardless of the weight precision, 2, 3, or 4 bits. However, the `C' networks with the conventional retraining show a better result when the weight precision is 4bits.  Overall, the proposed method yields improved performances.  

\subsection{Language modeling using recurrent neural networks}
\label{ssec:rnn}

Character-level language modeling predicts the next character, and is used for speech recognition and text generation. Since the input and output layers consider only alphabets, the input and output complexity are much lower than the word level language model. We adopt English Wikipedia dataset for training the character level language modeling. The dataset contains 100 MB English Wikipedia text. The input and output layers are composed of 256 units for one-hot encoded ASCII code. The RNN consists of three Long Short-Term Memory (LSTM) layers with a different number of memory cells ranging from 64 to 256~\cite{gers2002learning}. We train the RNNs using AdaDelta based SGD with 64 parallel input streams. The networks are unrolled 256 times and weights update is performed for128 forward steps. The learning rate starts from 5e-4 and decreases until 5e-8. For the step size adaptation, `1 epoch update \& fix' strategy is employed. The fixed-point optimization results are reported in \tablename~\ref{table:rnn}. As with our previous FFDNN and CNN results, it shows much improved performances on low-precision weights or small size networks.

\section{Concluding remarks}
\label{sec:5}
We have developed improved fixed-point weight optimization methods for deep neural networks. The first one adaptively determines the quantization step size   by measuring the weight distribution during the retraining procedure. The second one is a curriculum style fixed-point optimization technique, which conducts fixed-point optimization from high- to low-precision gradually. The proposed work yields better quantization results in FFDNN, CNN, and RNN experiments. Especially the effectiveness of the proposed techniques increases when the number of quantization levels is small and the network size is not large enough.
%



\bibliographystyle{IEEEbib}
\bibliography{refs}

\begin{thebibliography}{10}

\bibitem{amodei2015deep}
Dario Amodei, Rishita Anubhai, Eric Battenberg, Carl Case, Jared Casper, Bryan
  Catanzaro, Jingdong Chen, Mike Chrzanowski, Adam Coates, Greg Diamos, et~al.,
\newblock ``Deep speech 2: End-to-end speech recognition in english and
  mandarin,''
\newblock in {\em ICML 2016: 33rd International Conf. Machine Learning}, 2016.

\bibitem{He_2016_CVPR}
Kaiming He, Xiangyu Zhang, Shaoqing Ren, and Jian Sun,
\newblock ``Deep residual learning for image recognition,''
\newblock in {\em The IEEE Conference on Computer Vision and Pattern
  Recognition (CVPR)}, June 2016.

\bibitem{shin2016fixed}
Sungho Shin, Kyuyeon Hwang, and Wonyong Sung,
\newblock ``Fixed-point performance analysis of recurrent neural networks,''
\newblock in {\em 2016 IEEE International Conference on Acoustics, Speech and
  Signal Processing (ICASSP)}. IEEE, 2016, pp. 976--980.

\bibitem{hwang2014fixed}
Kyuyeon Hwang and Wonyong Sung,
\newblock ``Fixed-point feedforward deep neural network design using weights
  +1, 0, and -1,''
\newblock in {\em 2014 IEEE Workshop on Signal Processing Systems (SiPS)}.
  IEEE, 2014, pp. 1--6.

\bibitem{lin2015fixed}
Darryl~D Lin, Sachin~S Talathi, and V~Sreekanth Annapureddy,
\newblock ``Fixed point quantization of deep convolutional networks,''
\newblock in {\em ICML 2016: 33rd International Conf. Machine Learning}, 2016.

\bibitem{han2015deep}
Song Han, Huizi Mao, and William~J Dally,
\newblock ``Deep compression: Compressing deep neural network with pruning,
  trained quantization and huffman coding,''
\newblock {\em CoRR, abs/1510.00149}, vol. 2, 2015.

\bibitem{kim2014x1000}
Jonghong Kim, Kyuyeon Hwang, and Wonyong Sung,
\newblock ``X1000 real-time phoneme recognition vlsi using feed-forward deep
  neural networks,''
\newblock in {\em 2014 IEEE International Conference on Acoustics, Speech and
  Signal Processing (ICASSP)}. IEEE, 2014, pp. 7510--7514.

\bibitem{park2016fpga}
Jinhwan Park and Wonyong Sung,
\newblock ``{FPGA} based implementation of deep neural networks using on-chip
  memory only,''
\newblock in {\em 2016 IEEE International Conference on Acoustics, Speech and
  Signal Processing (ICASSP)}. IEEE, 2016, pp. 1011--1015.

\bibitem{HanLMPPHD16}
Song Han, Xingyu Liu, Huizi Mao, Jing Pu, Ardavan Pedram, Mark~A. Horowitz, and
  William~J. Dally,
\newblock ``{EIE:} efficient inference engine on compressed deep neural
  network,''
\newblock in {\em 43rd {ACM/IEEE} Annual International Symposium on Computer
  Architecture, {ISCA} 2016, Seoul, South Korea, June 18-22, 2016}, 2016, pp.
  243--254.

\bibitem{lee2016fpga}
Minjae Lee, Kyuyeon Hwang, Jinhwan Park, Choi Sungwook, Shin Sungho, and
  Wonyong Sung,
\newblock ``{FPGA}-based low-power speech recognition with recurrent neural
  networks,''
\newblock in {\em 2016 IEEE Workshop on Signal Processing Systems (SiPS)}.
  IEEE, 2016.

\bibitem{fraser2017scaling}
Nicholas~J Fraser, Yaman Umuroglu, Giulio Gambardella, Michaela Blott, Philip
  Leong, Magnus Jahre, and Kees Vissers,
\newblock ``Scaling binarized neural networks on reconfigurable logic,''
\newblock in {\em To appear in the PARMA-DITAM workshop at HiPEAC}, 2017, vol.
  2017.

\bibitem{anwar2015fixed}
Sajid Anwar, Kyuyeon Hwang, and Wonyong Sung,
\newblock ``Fixed point optimization of deep convolutional neural networks for
  object recognition,''
\newblock in {\em 2015 IEEE International Conference on Acoustics, Speech and
  Signal Processing (ICASSP)}. IEEE, 2015, pp. 1131--1135.

\bibitem{bengio2009curriculum}
Yoshua Bengio, J{\'e}r{\^o}me Louradour, Ronan Collobert, and Jason Weston,
\newblock ``Curriculum learning,''
\newblock in {\em Proceedings of the 26th annual international conference on
  machine learning}. ACM, 2009, pp. 41--48.

\bibitem{garofolo1993darpa}
John~S Garofolo, Lori~F Lamel, William~M Fisher, Jonathon~G Fiscus, and David~S
  Pallett,
\newblock ``{DARPA} {TIMIT} acoustic-phonetic continous speech corpus cd-rom.
  nist speech disc 1-1.1,''
\newblock {\em NASA STI/Recon technical report n}, vol. 93, 1993.

\bibitem{graves2013speech}
Alex Graves, Abdel-rahman Mohamed, and Geoffrey Hinton,
\newblock ``Speech recognition with deep recurrent neural networks,''
\newblock in {\em 2013 IEEE international conference on acoustics, speech and
  signal processing}. IEEE, 2013, pp. 6645--6649.

\bibitem{sutskever2013importance}
Ilya Sutskever, James Martens, George~E Dahl, and Geoffrey~E Hinton,
\newblock ``On the importance of initialization and momentum in deep
  learning.,''
\newblock {\em ICML (3)}, vol. 28, pp. 1139--1147, 2013.

\bibitem{43442}
Sergey Ioffe and Christian Szegedy,
\newblock ``Batch normalization: Accelerating deep network training by reducing
  internal covariate shift,''
\newblock in {\em ICML}, 2015, pp. 448--456.

\bibitem{sung2015resiliency}
Wonyong Sung, Sungho Shin, and Kyuyeon Hwang,
\newblock ``Resiliency of deep neural networks under quantization,''
\newblock {\em arXiv preprint arXiv:1511.06488}, 2015.

\bibitem{netzer2011reading}
Yuval Netzer, Tao Wang, Adam Coates, Alessandro Bissacco, Bo~Wu, and Andrew~Y
  Ng,
\newblock ``Reading digits in natural images with unsupervised feature
  learning,''
\newblock 2011.

\bibitem{sermanet2012convolutional}
Pierre Sermanet, Soumith Chintala, and Yann LeCun,
\newblock ``Convolutional neural networks applied to house numbers digit
  classification,''
\newblock in {\em Pattern Recognition (ICPR), 2012 21st International
  Conference on}. IEEE, 2012, pp. 3288--3291.

\bibitem{lecun1989backpropagation}
Yann LeCun, Bernhard Boser, John~S Denker, Donnie Henderson, Richard~E Howard,
  Wayne Hubbard, and Lawrence~D Jackel,
\newblock ``Backpropagation applied to handwritten zip code recognition,''
\newblock {\em Neural computation}, vol. 1, no. 4, pp. 541--551, 1989.

\bibitem{krizhevsky2012cuda}
Alex Krizhevsky,
\newblock ``cuda-convnet: High-performance c++/cuda implementation of
  convolutional neural networks,'' 2012.

\bibitem{courbariaux2015binaryconnect}
Matthieu Courbariaux, Yoshua Bengio, and Jean-Pierre David,
\newblock ``Binaryconnect: Training deep neural networks with binary weights
  during propagations,''
\newblock in {\em Advances in Neural Information Processing Systems}, 2015, pp.
  3123--3131.

\bibitem{gers2002learning}
Felix~A Gers, Nicol~N Schraudolph, and J{\"u}rgen Schmidhuber,
\newblock ``Learning precise timing with lstm recurrent networks,''
\newblock {\em Journal of machine learning research}, vol. 3, no. Aug, pp.
  115--143, 2002.

\end{thebibliography}

\end{document}